%% file: root.tex
\documentclass[letterpaper, 10pt, conference]{ieeeconf} 
 
\IEEEoverridecommandlockouts
\usepackage[pdftex]{graphicx}
\graphicspath{{figures/}}
\usepackage{url}
\usepackage{footmisc}
\usepackage{hyperref}
\hypersetup{
	colorlinks=true,
	linkcolor=black,
	citecolor=black,
	filecolor=black,
	urlcolor=black,
}
\usepackage{mathrsfs}
\usepackage[T1]{fontenc}
\usepackage[utf8]{inputenc}
\usepackage{csquotes}
\usepackage[english]{babel}
\usepackage[export]{adjustbox}
\usepackage{caption}
\usepackage{subcaption}
\usepackage[colorinlistoftodos,backgroundcolor=orange!20,bordercolor=orange]{todonotes}
\usepackage{lipsum}
\usepackage{dsfont}
\usepackage[letterpaper,left=.75in,right=.75in,top=.72in,bottom=.8in]{geometry}
\usepackage{amsmath}
\usepackage{amssymb}
\usepackage{mathtools}
\usepackage{calc}
\usepackage{pgfplots}
\usepackage{siunitx}
\usepackage{cases}
\usepackage{tabularx,booktabs}
\newcolumntype{C}{>{\centering\arraybackslash}X} 
\usepackage{placeins}
\usepackage{multirow}

\usepackage{todonotes}
\usepackage{xcolor,colortbl}
\usepackage{float}
\usepackage{subcaption}

\usepackage[style=ieee,
            doi=false,
            url=false,
            mincitenames=1,
            maxcitenames=1,
            minbibnames=6,
            maxbibnames=6,
            backend=biber]{biblatex}  
\title{%
\LARGE \bf Improving Lidar-Based Semantic Segmentation of Top-View Grid Maps by Learning Features in Complementary Representations}

\author{Frank Bieder$^{1,2,\ast}$, Maximilian Link$^{3}$, Simon Romanski$^{3}$, Haohao Hu$^{2}$   and
		Christoph Stiller$^{1,2}$%
    \thanks{$^{1}$ Authors are with the Mobile Perception Systems Department, FZI Research Center for Information Technology, Karlsruhe, Germany
    {\tt\small \{bieder, stiller\}@fzi.de}}%
\thanks{$^{2}$ Authors are with the Institute of Measurement and Control Systems, Karlsruhe Institute of Technology (KIT), Karlsruhe, Germany.}
\thanks{$^{3}$ Authors are now with understand.ai, 76227 Karlsruhe. They contributed to this work as part of their research stays at the Institute of Measurement and Control Systems, Karlsruhe, Germany.}
\thanks{$^{\ast}$Corresponding author.}
}

\begin{document}
\maketitle

\pubid{\begin{minipage}{\textwidth}~\\[12pt] \centering%
  \copyright~2020 IEEE. Personal use of this material is permitted. Permission from IEEE must be obtained for all other uses, in any current or future media, including reprinting/republishing this material for advertising or promotional purposes, creating new collective works, for resale or redistribution to servers or lists, or reuse of any copyrighted component of this work in other works.
\end{minipage}}
\pubidadjcol

\definecolor{vehicle}{rgb}{0.0,0.0,1.0}
\definecolor{two-wheel}{rgb}{0.11764705882352941,0.23529411764705882,0.5882352941176471}
\definecolor{pedestrian}{rgb}{1.0,0.11764705882352941,0.11764705882352941}
\definecolor{road}{rgb}{1.0,0.0,1.0}
\definecolor{sidewalk}{rgb}{0.29411764705882354,0.0,0.29411764705882354}
\definecolor{parking+other-ground}{rgb}{1.0,0.5882352941176471,1.0}
\definecolor{building+fence}{rgb}{1.0,0.7843137254901961,0.0}
\definecolor{pole+traffic-sign}{rgb}{1.0,0.47058823529411764,0.19607843137254902}
\definecolor{vegetation}{rgb}{0.0,0.6862745098039216,0.0}
\definecolor{trunk}{rgb}{0.5294117647058824,0.23529411764705882,0.0}
\definecolor{terrain}{rgb}{0.5882352941176471,0.9411764705882353,0.3137254901960784}

\pagestyle{empty}

\input{content/01_abstract}

\input{content/02_introduction}
\input{content/03_main}

\input{content/04_results_and_evaluation}

\input{content/05_conclusions}

\printbibliography

\end{document}

%% file: content/01_abstract.tex
\begin{abstract}
    In this paper we introduce a novel way to predict semantic information from sparse, single-shot LiDAR measurements in the context of autonomous driving. In particular, we fuse learned features from complementary representations. The approach is aimed specifically at improving the semantic segmentation of top-view grid maps. Towards this goal the 3D LiDAR point cloud is projected onto two orthogonal 2D representations. For each representation a tailored deep learning architecture is developed to effectively extract semantic information which are fused by a superordinate deep neural network. The contribution of this work is threefold: (1) We examine different stages within the segmentation network for fusion. (2) We quantify the impact of embedding different features. (3) We use the findings of this survey to design a tailored deep neural network architecture leveraging respective advantages of different representations. Our method is evaluated using the SemanticKITTI dataset which provides a point-wise semantic annotation of more than 23.000 LiDAR measurements.
    
    \textit{Index Terms}--- Semantic Segmentation, Sensor Fusion, Multi-Representation Encoding, Deep Learning, Automated Driving
    \end{abstract}

%% file: content/02_introduction.tex
\section{Introduction}

Developing robust and accurate scene understanding is crucial towards the goal of deploying and integrating autonomous agents in society. A key component of such systems is a reliable semantic segmentation of the agent's environment.

An autonomous agent draws from a wide range of sensor data differing in its characteristics. While LiDAR point clouds provide precise spatio-temporal information, obtaining semantic information through deep learning based methods is challenging due to the sparsity of the data. Combining a reliable semantic segmentation with the spatial precision of the point cloud inherently yields a strong environment representation.

\begin{figure}[!t]
    \vspace*{2mm}
    \centering
    \includegraphics[width =0.90\linewidth]{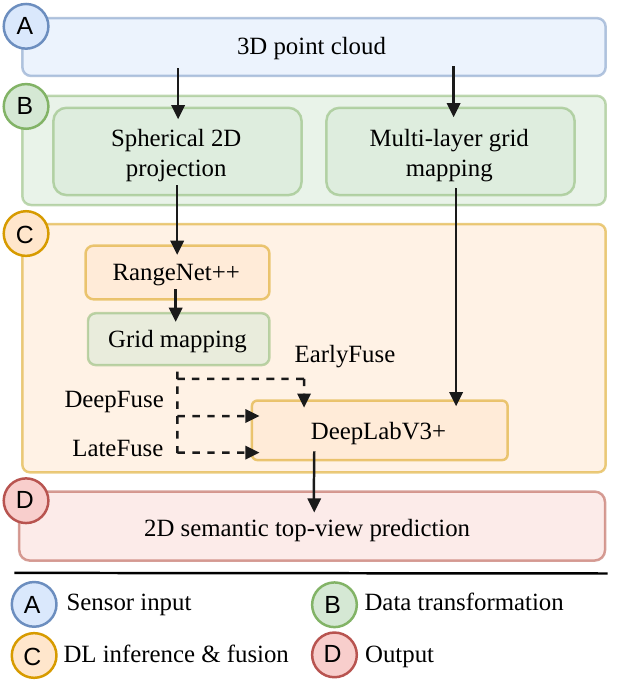}
    \caption{
        System Overview of our Fusion Pipeline. We encode a 3D point cloud into a spherical 2D projection and a multi-layer grid map. We segment the spherical projection and project the results into a grid map representation for fusion. To obtain a 2D semantic top view prediction, we explore fusing the grid map information and the intermediate segmentation results at three different stages using DeeplabV3.
    } 
    \label{fig:overview}
    \end{figure}

Projecting the 3D point clouds into lower-dimensional representations creates density which facilitates the semantic segmentation but causes loss of spatial information. Of the commonly used representations range images and grid maps are of interest as they represent orthogonal projections of the point cloud and could thus be used as efficient source for obtaining semantic information through segmentation networks. Their orthogonality allows neural networks to specialize on complementary object geometries and thus different semantic classes.

\pubidadjcol
With this work we are attempting to improve the efficiency of environment perception for autonomous agents by using different environment representations from a single sensor source at a single point in time. By doing so we can both gain redundancy in the scene classification for which otherwise additional sensors would be required and improve the overall segmentation results. The later stems from the segmentation sensitivity of the segmentation networks towards the respective representation. Beyond the benefit of using complementary environment representations, our approach is able to predict a dense semantic environment based on a sparse point cloud. The overview of our approach is presented in Figure \ref{fig:overview}.

\textit{Statement of Contribution:}
We simultaneously learn semantic segmentation of sparse LiDAR measurements in different representations. We propose multiple fusion strategies to fuse all predicted information into top view representation. In addition, we vary the encoding of features which are transformed into target representation in richness and nature to evaluate their effectiveness for the segmentation task. We show that our approach is improving the semantic grid map segmentation and significantly outperforms the baselines.


\section{Related Work}
\subsection{Environment Representations} 
A key differentiator between representations is the partition strategy of three-dimensional environment.

Using point clouds directly is not practical due to the large amount of data per scan with a lack of structure rendering many segmentation approaches unsuitable and prohibitively expensive.
Voxels partition the environment into a 3D grid of cubes of equal size \cite{Zhou2018}, \cite{Wang2015} and often suffer from the curse of dimensionality.

Stixels \cite{Pfeiffer2011}, \cite{Cordts2017} have a square base on planar ground and are defined by its position and height. 
These partition strategies share the disadvantage with graph-based approaches such as Octrees \cite{Ohtake2003} that fully convolutional segmentation approaches cannot be applied to their structure.

Grid maps are an orthographic top view where the environment is divided into equidistant cells. Occupancy grid maps \cite{Elfes1987}, \cite{Moravec1989} encode if a cell is occupied and gave rise to multi-layer grid maps \cite{Wirges2018Obj, Wirges2019}. The latter provide efficient, scale invariant frames for sensor fusion.

Spherical Projections such as Range Images \cite{Milioto2019} project range and intensity measurements from a point cloud onto a planar image. Their orthogonality to grid maps makes their segmentation sensitivities largely complementary.

\label{sec:related:pointcloud}
\subsection{Point Cloud Segmentation} \textit{Guo et. al} \cite{Guo2020} define different strategies based on paradigms: 
Projection based methods such as SqueezeSeg \cite{Wu2019} project 3D point clouds onto 2D images and apply convolutional neural networks for segmentation. For this work we choose two different projection based segmentation methods, namely Rangenet++ \cite{Milioto2019} and the work proposed in \textit{Bieder et. al.} \cite{Bieder2020SemGridEsti} based on Deeplab \cite{Chen2018}. 
Discretization based methods \cite{Jing2016} apply a discrete lattice on the point cloud which is fed into 3D convolutions. The maximum segmentation performance is directly affected by the granularity of the discretization. 
Point based methods such as \cite{Hu2020}, \cite{Qi2017a} and \cite{Qi2017b} mostly use multi-layer perceptrons with shared weights to learn per-point features. The problem here is that using this approach the geometric relations among the points is disregarded. Different approaches exist to tackle this problem, trying to incorporate features from neighboring points. Hybrid methods form the last paradigm but are irrelevant for this work.

\subsection{Deep Learning Based Image Segmentation} 
Image segmentation techniques \cite{Krizhevsky2017} can be transferred to the field of semantic segmentation: \textit{Hariharan et. al} \cite{Hariharan2014} and \textit{Girshick et. al.} \cite{Girshick2014} employ algorithms to get region proposals, then use deep convolutional neural networks (DCNNs) to extract features for these regions and then perform classification based on these features. \textit{Farabet et. al.} \cite{Farabet2013} use DCNNs on the whole image to extract features on different scales but segment using classifiers.

The current predominant strategy for image segmentation is using fully convolutional networks applying a DCNN to the whole image such as in \cite{Long2015}. Recent frameworks \cite{Chen2018} achieve a pixel-wise segmentation by leveraging resizing, skip connections and atrous convolutions to increase the resolution after downsampling with strided convolutions. 

\label{sec:fusion-strategies}
\subsection{Fusion Strategies} \textit{Eitel et. al.} \cite{Eitel2015} and \textit{Hazirbas et. al.} \cite{Hazirbas2017} introduce and explore early and late fusion strategies of RGB and depth information for semantic labeling of indoor scenes and object detection. \textit{Richter et. al.} \cite{Richter2019} fuse range measurements and semantic estimates in through a evidential framework. \textit{Zhang et. al.} \cite{Zhang2015} apply different fusion concepts for separate already segmented  LiDAR and Camera data.

%% file: content/03_main.tex

\section{Methodology}

Our goal is to predict a semantic top-view grid map based on a single-shot LiDAR measurement. We encode a 3D point cloud into a spherical 2D projection, denoted as range image, and a top view multi-layer grid map consisting of five layer. The former is processed by a Rangenet++ (RNet), trained on SemanticKITTI to predict eleven semantic classes. Subsequently, the features predicted by RNet are also projected into the grid map representation. Our goal is to fuse both feature sources within a deep learning architecture to improve the semantic grid map estimation. This challenge requires multiple design choices: We explore fusing the heterogenous features at three different stages of the DeeplabV3 (DLV3) architecture. In addition, we use four different encodings for the RNet output within the fusion process. In contrast, the encoding of the multi-layer grid maps is kept constant through all experiments.

\begin{figure}[ht]
    \vspace*{2mm}
    \centering
    \includegraphics[width=0.48\textwidth]{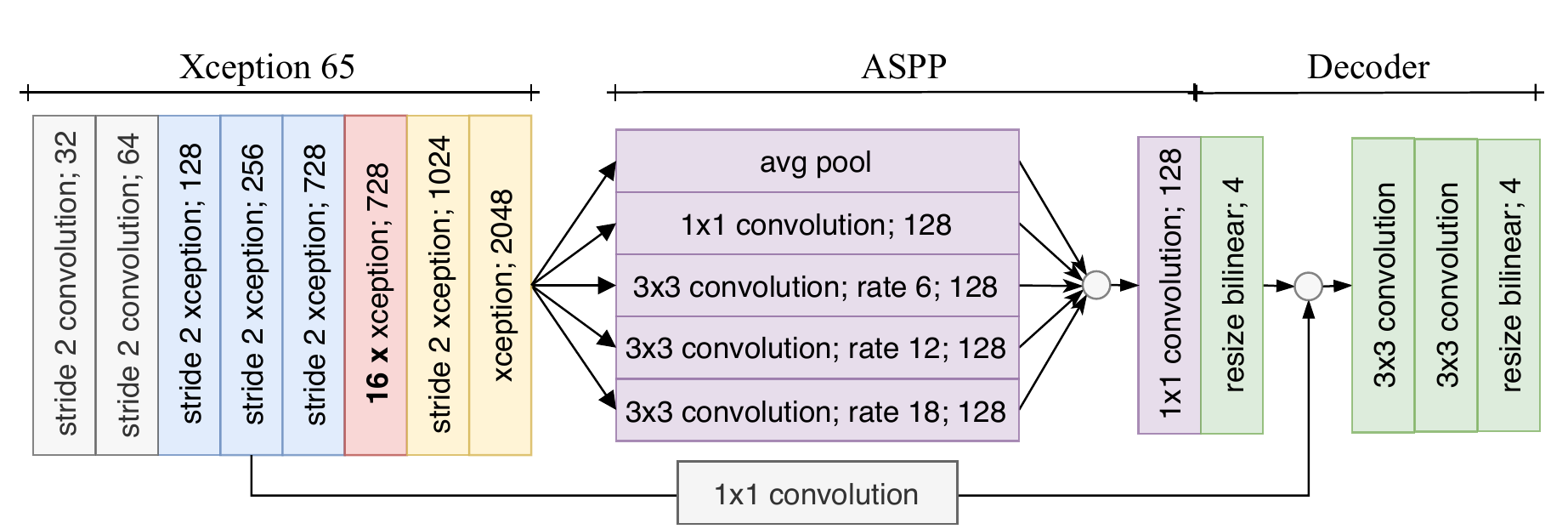}
\caption{DeepLabV3 (DLV3) Architecture. Each xception block is defined by a stride factor and its number of filters. For information regarding \textit{Xception65}, \textit{ASPP} or its \textit{Decoder} see \cite{Chen2018}.} 
\label{fig:deeplab}
\end{figure}

\begin{figure*}[thbp]
    \vspace*{2mm}
    \centering
    \includegraphics[angle = 270, width=\textwidth]{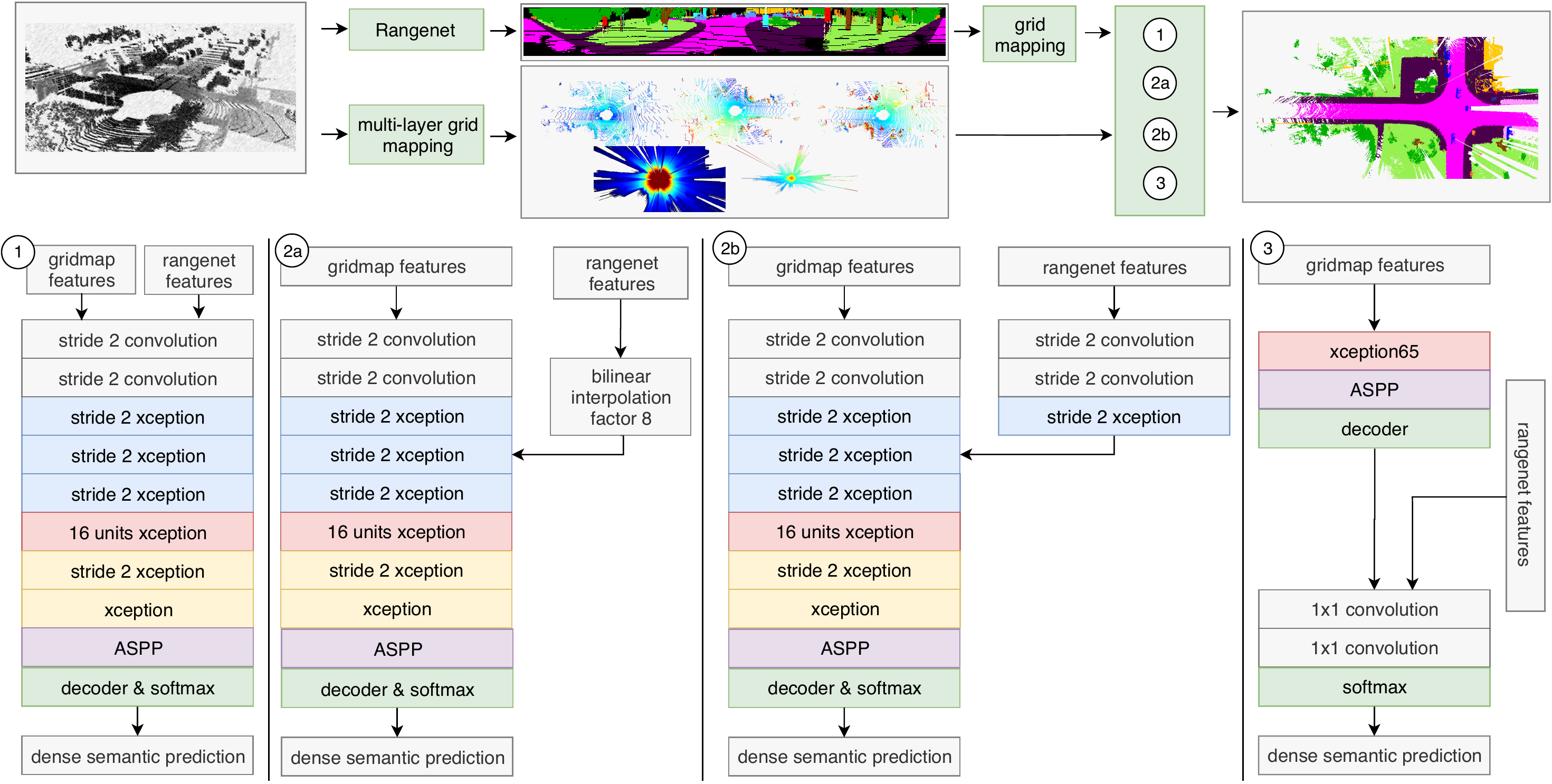}
    \vspace*{0.2cm}
    \caption{
    Details of our Deeplab Architecture Variations. (1) EarlyFuse: Early fusion with concatenated input, (2a) DeepFuseA: Deep fusion with bilinear interpolation, (2b) DeepFuseB: Deep fusion with block duplication, (3) LateFuse: Late fusion with neural fusion. In this visualization the four different feature variations resulting from the grid mapped RNet output are not further displayed but descrped in . 
    }
    \label{fig:combined_flow}
\end{figure*}

\subsection{Feature Encoding of Multi-Layer Grid Maps}
Here, it is described how we encode a raw LiDAR measurement $P$, consisting of a 3D point cloud with intensity values, into a rich multi-layer grid map representation, later denoted as $f_{\text{GM}}$. It consists of five individual layers $f_{\text{GM}} = \{f_{z+},f_{z-},f_{i},f_{\#},f_{\overline{z}}\}$, which are defined as follows:

\begin{enumerate}
    \item \textit{Detections Z Max} ($f_{z+}$): The maximum observed height of any detection in the cell.
    \begin{equation}
        f_{z+} = \max_{p \in P_C}\{z(p)\}
    \end{equation}
    where $p$ denotes a single point of the set of points $P_C$ within one grid cell $C$. $z(p)$ defines the height value and $i(p)$ the intensity value of point $p$.
    \item \textit{Detections Z Min} ($f_{z-}$): The minimum observed height of any detection in the cell.
    \begin{equation}
        f_{z-} = \min_{p \in P_C}\{z(p)\}
    \end{equation}
    \item \textit{Intensity} ($f_{i}$): The mean intensity of all detections in the cell.
    \begin{equation}
        f_{i} = \frac{1}{|P_C|}\sum_{p \in P_C} i(p)
    \end{equation}
    \item \textit{Observations} ($f_{\#}$): The number of LiDAR rays that have crossed the cell. Each point $p$ has a ray $\mathrm{ray}(p)$ associated with it.
    \begin{equation}
        f_{\#}=|P_{\mathrm{ray}}|
    \end{equation}
    with $P_{\mathrm{ray}} = \{p \in P : \mathrm{ray}(p) \dashv C \}$ where $\dashv$ denotes the intersection of a ray with grid cell $C$. 
    \item \textit{Occlusions Upper Bound} ($f_{\overline{z}}$): The maximum possible height for objects in the cell given by the lowest ray that crossed that cell.
    \begin{equation}
        f_{\overline{z}}=\min_{p_{\mathrm{ray}} \in P_{\mathrm{ray}}}\{z(p_{\mathrm{ray}})\}
    \end{equation}
\end{enumerate}
To generate the two semi-dense layers, $f_{\#}$ and $f_{\overline{z}}$, we map the point cloud in polar coordinates. The sparse layers are directly generated in cartesian coordinates. Our first baseline experiment \textit{GMBase} is conducted by training solely on $f_{\text{GM}}$ and using the DLV3+ architecture as it represents the best performing experiment in \cite{Bieder2020SemGridEsti}.

\begin{table*}[!t]
    \vspace*{2mm}
    \centering
     \caption{Selection of Quantitative Results for Sparse and Dense Segmentation Tasks. $\mathrm{IoU_{c}}$ and $\mathrm{mIoU}$ are reported in percent. All experiments except RNBase make use of $f_{\text{GM}}$}\label{tab::exp_results}
    \label{tab:results1}
    \begin{tabular}{llr|*{11}{p{0.70cm}}cc}
    \toprule
    &Approach &\textbf{mIoU}&build-ing&park-ing&pedest-rian&pole&road&side-walk&terrain&trunk&two wheel&vege-tation&vehicle\\
    \midrule
    \multirow{7}{*}{\rotatebox{90}{Sparse}}
    & GMBase, only $f_{\text{GM}}$ & \textbf{45.32} & 66.37 & 18.58 & 0 & 25.18 & 87.91 & 63.2 & 68.52 & 14.7 & 4.24 & 74.78 & 75.02  \\
    & RNBase, no $f_{\text{GM}}$ & \textbf{50.44} & 59.63 & 38.14 & 18.09 & 23.1 & 91.52 & 73.82 & 70.65 & 23.49 & 21.9 & 71.3 & 63.16 \\
    & EarlyFuse-$f_{H}$ & \textbf{56.46} & 72.83 & 34.69 & 18.88 & 41.61 & 90.84 & 73.1 & 71 & 36.66 & 24.14 & 76.15 & 81.17 \\
    & EarlyFuse-$f_\Sigma$ & \textbf{56.35} & 72.35 & 35.38 & 20.36 & 41.36 & 90.9 & 73.07 & 71.08 &   37.81 & 20.68 & 75.96 & 80.86 \\
    & EarlyFuse-$f_p$ & \textbf{53.32} & 73 & 34.46 & 11.88 & 33.2 & 90.54 & 72.67 & 71.57 & 28.91 & 12.27 & 76.45 & 81.55 \\
    & DeepFuseB-$f_\Sigma$ & \textbf{51.44} & 69.76 & 31.15 & 0.63 & 33.18 & 90.3 & 72.04 & 70.53 & 31.7 & 12.69 & 75.48 & 78.41 \\
    & LateFuse-$f_\Sigma$ & \textbf{48.83} & 53.76 & 35.87 & 14.92 & 26.35 & 86.22 & 69.33 & 66.52 & 26.6 & 22.19 & 68.68 & 66.74 \\
    \addlinespace
    \multirow{7}{*}{\rotatebox{90}{Dense}}
    & GMBase, only $f_{\text{GM}}$ & \textbf{32.72} & 46.42 & 15.95 & 0 & 15.25 & 81.71 & 44.82 & 56.6 & 6.92 & 1.38 & 52.84 & 38.0 \\
    & RNBase, no $f_{\text{GM}}$ & \textbf{13.36} & 15.58 & 8.44 & 6.76 & 16.78 & 19.03 & 16.82 & 14.66 & 12.87 & 10.14 & 23.07 & 2.79 \\
    & EarlyFuse-$f_{H}$ & \textbf{35.6} & 44.59 & 18.8 & 4.85 &   23.38 & 79.9 & 45.64 & 54.7 & 18.28 & 8.59 & 51.97 & 40.9 \\
    & EarlyFuse-$f_\Sigma$ & \textbf{35.9} & 48.06 & 20.48 & 4.15 & 23.07 &  80.46 & 46.3 & 55.22 &  17.49 & 6.92 & 52.56 & 40.19 \\
    & EarlyFuse-$f_p$ & \textbf{34.4} & 49.27 & 20.89 & 3.04 & 18.92 & 80.92 & 45.36 &  53.52 & 14.06 & 4.17 & 49.67 & 38.58 \\
    & DeepFuseB-$f_\Sigma$ & \textbf{33.37} & 47.18 & 19.37 & 0.15 & 19.16 &  78.16 & 45.27 & 53.47 & 14.96 & 4.52 & 52.08 & 32.77 \\
    & LateFuse-$f_\Sigma$ & \textbf{15.38} & 11.5 & 7.54 & 4.35 & 17.75 &   24.92 & 19.95 & 14 & 13.1 & 9.46 & 27 & 19.65 \\
    \bottomrule
    \end{tabular}
    \end{table*}

\subsection{Feature Encoding of RNet++ Output}
We extract semantic pseudo probabilities for each point in the point cloud by applying RNet. Subsequently, these information are transformed into a top-view grid map representation. Multiple points can be assigned to one grid cell and we propose several ways to aggregate and encode the semantic information for each grid cell. This allows us to craft a variety of features:
\begin{enumerate}
    \item \textit{Histogram of Predictions} ($f_{H}$): A histogram $f_{H} = [c_1, ..., c_n]$ is defined for each cell $C$. There are as many layers as classes $n$ and each layer stores the number of points that were classified by RNet and associated with cell cell $C$.
    \begin{equation}
        c_i = \sum_{p \in P_C}^{n} {1}_{[c_i}(p)
    \end{equation}
    \item \textit{Argmax} ($f_p$): It is defined in cell $C$. This requires only one layer, the classes are encoded and the cell gets assigned the code of the class of which it contains most points. This is equivalent to the argmax of the histogram feature.
    \begin{equation}
        f_p = \arg\max f_{H}
    \end{equation}
    \item \textit{Summed Probabilities} $f_\Sigma$): It is defined for each class in a given cell. Like above we have as many layers as classes. But for this feature we sum the probabilities for the given class across all points in the cell. For one class $c$ the feature $v_{c}$ in a cell $C$ can be formalized as
    \begin{equation}
        v_{c} = \sum_{p \in P_C}^{n}P_{r}(c \: | \: p)
    \end{equation} where $P_{r}(c \: | \: p)$ is the softmax outputs of RNet representing the probabilities for a point $p$ belonging to a class $c$.
    \item \textit{Mean Probabilities} ($f_\mu$): It is defined for each class in a given cell. We have as many layers as classes but we store the mean softmax activation instead of the sum. Following above notation the feature can be obtained from
    \begin{equation}
        v_{c} = \frac{1}{|P_C|} \sum_{p \in P_C}^{n}P_{r}(c \: | \: p).
    \end{equation}
\end{enumerate}

In case the feature encoding is not chosen or specified we denote the RNet features $f_{R}$. As our experiments have shown that $f_\mu$ is continuously outperformed by the similar encoding $f_\Sigma$, we only report experiments with the later. An experiment solely based on the RNet predictions defines our second baseline \textit{RNBase}.

\subsection{Fusion Strategies}
Given features from two sources we propose several fusion strategies, which are presented in Figure 2 and described in the following:

\subsubsection[Early]{Early Fusion (EarlyFuse)}
We directly concatenate the layers encoding the RNet output and the multi-layer grid map $f_{\text{GM}}$ and feed the resulting tensor in the DLV3 model. This allows DLV3 to combine features from both sources right at the beginning. The drawback of this approach is that the semantic features are not extracted in parallel but in series which might lead to a longer runtime and is not optimal for autonomous systems.
\subsubsection[Deep]{Deep Fusion (DeepFuse)}

All types of RNet features $f_R$ can be fed into the feature extractor in an intermediate layer of the  DLV3 model. This necessitates the extension of the DLV3 network by one branch that converts the features to the required spatial resolution and feeds them to the correct hidden layer. We use two different approaches:
(a) We sample the feature down to the right resolution using bilinear interpolation. This approach is faster as the RNet network can run in parallel to the first layers of the DVL3 network. Deep fusion hence allows for a trade-off between runtime and segmentation performance. (b) We duplicate the normal blocks in the layer where we want to fuse them to allow it to do the same computation.

\begin{figure*}[!htbp]
    \vspace*{2mm}
    \renewcommand{\thesubfigure}{\arabic{subfigure}}
    \centering
    \begin{subfigure}{\textwidth}
    \parbox[c]{.03\linewidth}{\subcaption{}}
    \parbox[c]{\linewidth}{
        \includegraphics[angle=270, width = 0.95\textwidth ]{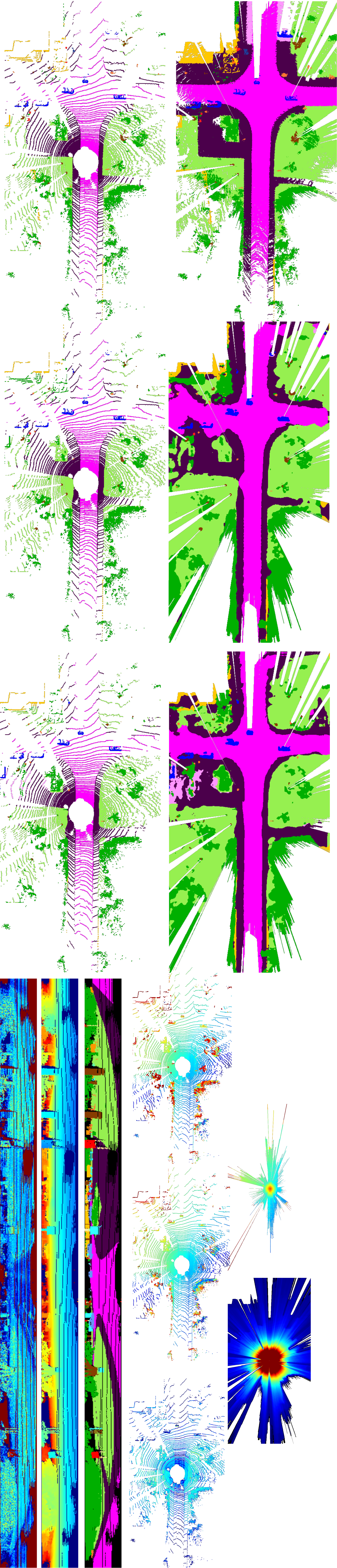}
    }
    \end{subfigure}
    \\ \vspace*{0.20cm}
    \begin{subfigure}{\textwidth}
        \parbox[c]{.03\linewidth}{\subcaption{}}
        \parbox[c]{\linewidth}{
            \includegraphics[angle=270, width = 0.95\textwidth ]{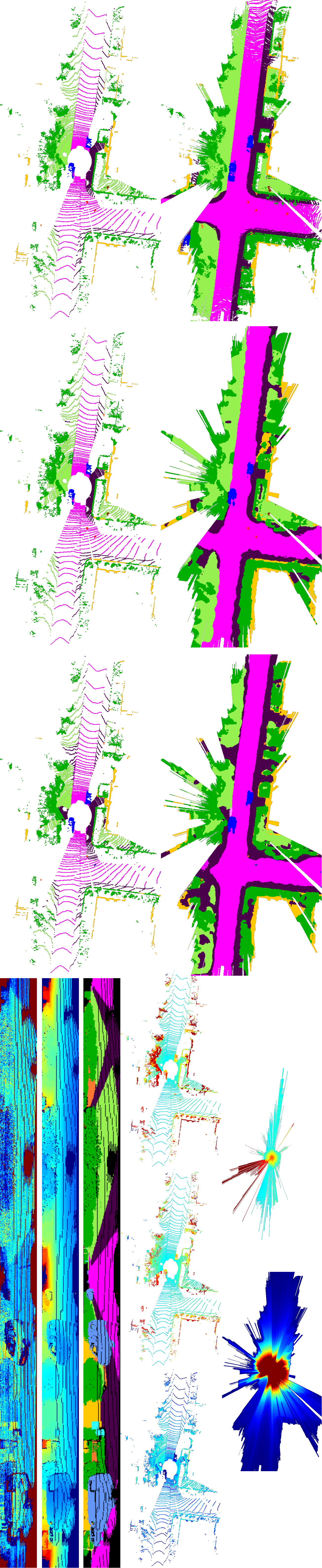}
        }
        \end{subfigure}
        \\ \vspace*{0.40cm}
        \begin{subfigure}{\textwidth}
            \centering
            \includegraphics[angle=270, width = 0.95\textwidth]{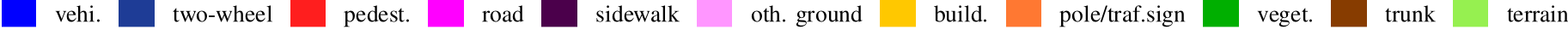}
          \end{subfigure}
        \\ \vspace*{0.20cm}
    \centering
    
    \caption{
        Qualitative Analysis of our Segmentation Results for Two Different Scans, S8-0000 (1) and S8-1531 (2). The figure is separated in four columns. The very left one visualizes a raw point cloud and its encoded features. The three range images represent the intensity and the distance of the point cloud and its inference made by a RNet. Below that the five multi-layer grid maps are shown (from l. to r. and t. to b.: $f_{i},f_{z-},f_{z+},f_{\#},f_{\overline{z}}$). The second column presents the results of the baseline GMBase and the third column our results of EarlyFuse-$f_{H}$. The last column presents the corresponding ground truth.
    }
    \label{fig:qualitative_results}
\end{figure*}

\subsubsection[Late]{Late Fusion (LateFuse)}
 
The RNet information can be fed into the softmax layer of DLV3. Here, $f_R$ and the corresponding feature layer within DLV3 have the same resolution and can be concatenated. Subsequently, the feature maps are combined using two 1$\times$1 convolutions forming new logits. 

Fusion at this stage is most efficient since the computationally expensive neural networks run in parallel and only a light combination step is performed afterwards. Leveraging additional information across grid map cells is impossible for fusion at this stage which is limiting the theoretically achievable segmentation performance.

\subsection{Ground Truth Generation and Learning Task}
We define two tasks for the evaluation of our experiments:
\begin{enumerate}
    \item \textit{Sparse Task}: The ground truth for this task is generated by processing a single annotated point cloud for each measurement. Hence, only gird cells with at least one LiDAR point can be taken for evaluation.
    \item \textit{Dense Task}: To generated the ground truth for this task we make use of the highly accurate SLAM poses of SemanticKITTI and accrete for each measurement the semantic information of all surrounding point clouds. The aggregated semantic information are transformed in the coordinate system of the current pose and subsequently mapped into a dense semantic top-view layer. Points from surrounding time stamps which carry information about dynamic object classes are rejected so that movable objects do not leave smears.
\end{enumerate}
Grid cells without assigned LiDAR points in the ground truth do not influence the evaluation result.

\subsection{Experimental Setup}
We reduce the SemanticKITTI label classes to eleven classes as we join moving classes with their non-moving counterpart and follow \cite{Bieder2020SemGridEsti} in aggregating classes with similar appearance. The unlabeled class is ignored.

For the training process we select all publicly available sequences of SemanticKITTI except for sequence S8 which is solely used for evaluation and consists of more than 4000 measurements. We flip the data randomly along the vertical axis and scale and crop with a factor between 0.8 and 1.25 with steps of 0.1. The scaling and cropping range is chosen rather small since the variation in scale for objects in a grid map is limited in comparison, e.g., to projecting camera images to range images.

The models are initialized by a pre-training on ImageNet \cite{Krizhevsky2017}. For DeepFuseB we duplicate the weights for the additional branch in pre-training. In all experiments, the input consists of information from one single-shot LiDAR measurement and choose a grid map resolution of $1001 \times 501$ grid cells. Furthermore, we  train on a NVIDIA RTX 2080 Ti GPU with 11GB memory allowing a batch size of two. All DLV3 models are trained for $500.000$ iterations from which we select the best performing model.

%% file: content/04_results_and_evaluation.tex
\section{Evaluation}

\subsection{Evaluation Metric}
We use the commonly used Intersection over Union ($\mathrm{IoU}$) \cite{Everingham2014} metric. For a class $c$ the $\mathrm{IoU_{c}}$ is calculated as:
\begin{equation}
    \mathrm{IoU_{c}} = \frac{\text{TP}_{c}}{\text{TP}_{c} + \text{FP}_{c} + \text{FN}_{c}}
\end{equation}
where $\text{TP}_{c}$ are the true positives, $\text{FP}_{c}$ the false positives and $\text{FN}_{c}$ the false negatives for class $c$. We report the IoU for all of the 11 classes. Additionally we report the mean IoU which is calculated as follows:
\begin{equation}
  \mathrm{mIoU} = \frac{1}{|C|} \sum_{c \: \in \: C} \mathrm{IoU_{c}}
\end{equation}
where $C$ is the set of classes.

\subsection{Quantitative Evaluation}

The experiments assess the impact of the two main degrees of freedom we have in our design.
(1) Different fusion techniques: EarlyFuse, DeepFuse and LateFuse and (2) the encoding of features infered by RNet.

We benchmark the results with respect to two baselines: 
\begin{itemize}

\item GMBase, as it is implemented in \cite{Bieder2020SemGridEsti}.
\item RNBase, using the argmax prediction of \cite{Milioto2019} transformed in a top-view grid map.

\end{itemize}

The $\mathrm{mIoU}$ results from the sparse evaluation confirm our intuition: Features extracted from the range image representation hold valuable information and complement well the top-view  multi-layer grid maps. The result is a significant performance boost - in particular using the feature encodings $f_{H}$ and $f_\Sigma$. With the exception of one experiment, all fusion approaches outperform both baselines. EarlyFuse achieves the strongest performance among all fusion modes with the drawback that it requires the most computation time. A possible explanation is the potential of the network to combine the feature information at all stages and hence at different layers of abstraction. Depending on the application the performance gap between EarlyFuse and DeepFuse could be outweighed by the considerably shorter computation time for DeepFuse. While EarlyFuse and DeepFuse are consistently outperform the baseline approaches, LateFuse performs similar to them with respect to both evaluation tasks. Table \ref{tab:results1} shows that the best classification performance is achieved by using EarlyFuse-$f_{H}$ reaching 56.46\% mIoU.

Analogous to the sparse evaluation, EarlyFuse-$f_{H}$ and EarlyFuse-$f_\Sigma$ provide the strongest classification performance for the task of dense semantic grid map estimation. We argue that for the dense evaluation the grid map features contribute stronger toward segmentation performance. This is explainable through the grid map layers $f_{\overline{z}}$ and $f_{\#}$ which posses a semi-dense structure due to the polar coordinate mapping of the raw measurements. Accordingly, they are the only information source for cells without detections. RNBase struggles with the dense task as it is not designed to conduct a dense prediction.

In a larger parameter study, features encoded as $f_\Sigma$ performed best across various fusion stages. Consequently, we selected this feature as reference to compare different fusion types. Furthermore, DeepFuseB mostly outperformed DeepFuseA.

\subsection{Qualitative Evaluation}

We qualitatively compare the results of our top performing architecture EarlyFuse-$f_{H}$ and GMBase, the top performer of \cite{Bieder2020SemGridEsti}, by examining some scenes in more detail. Figure \ref{fig:qualitative_results} displays the results for two scans of the evaluation sequence, the scan S8-0000 (1) and S8-1531 (2). We observe that both approaches are able to capture the general structure of the scene by correctly predicting most parts of road, sidewalk vegetation and also more challenging classes like vehicles. 

\begin{figure}[ht]\label{fig:zoom}
  \vspace*{2mm}
  \begin{subfigure}{.15\textwidth}
      \centering
      \includegraphics[width = 0.95\textwidth ]{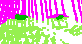} 
      \label{fig:sub-first}
    \end{subfigure}
    \begin{subfigure}{.15\textwidth}
      \centering
      \includegraphics[width = 0.95\textwidth]{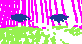}
      \label{fig:sub-second}
    \end{subfigure}
    \begin{subfigure}{.15\textwidth}
        \centering
        \includegraphics[width = 0.95\textwidth ]{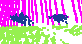}
        \label{fig:sub-third}
      \end{subfigure}\vspace*{0.3cm}\\

\begin{subfigure}{.15\textwidth}
\centering
\includegraphics[width = 0.95\textwidth ]{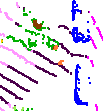} 
\label{fig:sub-first}
\end{subfigure}
\begin{subfigure}{.15\textwidth}
\centering
\includegraphics[width = 0.95\textwidth]{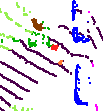}
\label{fig:sub-second}
\end{subfigure}
\begin{subfigure}{.15\textwidth}
  \centering
  \includegraphics[width = 0.95\textwidth ]{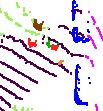}
  \label{fig:sub-third}
\end{subfigure}\\
\caption{Crop of two Instances of Two-Wheeler (top) and Pedestrians (bottom). Left: GMBase, middle: EarlyFuse-$f_{H}$, right: Ground truth.}
\label{fig:zoom} 
\end{figure}

While both approaches struggle at times to correctly classify small and distant objects, the misclassifications are much less pronounced in case of EarlyFuse-$f_{H}$. For instant, it correctly classifies two out of three pedestrians on the road of S8-1531, while non of them is detected by GMBase using solely multi-layer grid maps. Further examples of the superiority of our fusion approach are presented in Figure \ref{fig:zoom}. It shows two demanding classification scenes: One with two pedestrians and one with two two-wheeler instances which are encompassing motorcycles, bicycles and their potential riders. Again, GMBase fails to classify the given instances, while EarlyFuse-$f_{H}$ captures both two-wheelers and one pedestrian.


%% file: content/05_conclusions.tex
\section{Conclusions and Future Work}
\label{sec:conclusion}

In this work we show that fusing features learned in complementary representations significantly improves the semantic segmentation of top-view grid maps in the context of autonomous driving. 

We design different fusion architectures and feature encodings and examine their effectiveness given two segmentation tasks. The majority of our fusion strategies outperform all baseline approaches. Fusing representations at an early stage yields superior result in comparison to both previous work and other fusion strategies developed in this work. Driven by these results we propose a deep learning architecture that tailors the RangeNet++ and DeepLabV3+ networks and infers a rich and dense semantic top-view representation. 

Future research directions include using sequences of LiDAR point clouds to leverage temporal information for a segmentation that is consistent over time and the incorporation of additional heterogeneous sensor data such as radar and camera data.



%% file: content/references.bib
@inproceedings{Bieder2020SemGridEsti,
  author    = {{Bieder}, Frank and {Wirges}, Sascha and {Janosovits}, Johannes and {Richter}, Sven and {Wang}, Zheyuan and {Stiller}, Christoph},
  booktitle = {IEEE Intelligent Vehicles Symposium, Proceedings},
  title     = {{Exploiting Multi-Layer Grid Maps for Surround-View Semantic Segmentation of Sparse LiDAR Data}},
  year      = {2020}
}

@article{Chen2018,
  author   = {Chen, Liang Chieh and Papandreou, George and Kokkinos, Iasonas and Murphy, Kevin and Yuille, Alan L.},
  doi      = {10.1109/TPAMI.2017.2699184},
  eprint   = {1606.00915},
  journal  = {IEEE Transactions on Pattern Analysis and Machine Intelligence},
  title    = {{DeepLab: Semantic Image Segmentation with Deep Convolutional Nets, Atrous Convolution, and Fully Connected CRFs}},
  year     = {2018}
}

@inproceedings{Eitel2015,
  arxivid   = {1507.06821},
  author    = {Eitel, Andreas and Springenberg, Jost Tobias and Spinello, Luciano and Riedmiller, Martin and Burgard, Wolfram},
  booktitle = {IEEE International Conference on Intelligent Robots and Systems},
  title     = {{Multimodal deep learning for robust RGB-D object recognition}},
  year      = {2015}
}

@article{Everingham2014,
  author   = {Everingham, Mark and Eslami, S. M.Ali and {Van Gool}, Luc and Williams, Christopher K.I. and Winn, John and Zisserman, Andrew},
  journal  = {International Journal of Computer Vision},
  title    = {{The Pascal Visual Object Classes Challenge: A Retrospective}},
  year     = {2014}
}

@inproceedings{Hazirbas2017,
  author    = {Hazirbas, Caner and Ma, Lingni and Domokos, Csaba and Cremers, Daniel},
  booktitle = {Lecture Notes in Computer Science},
  title     = {{FuseNet: Incorporating depth into semantic segmentation via fusion-based CNN architecture}},
  year      = {2017}
}

@article{Krizhevsky2017,
  author  = {Krizhevsky, Alex and Sutskever, Ilya and Hinton, Geoffrey E.},
  journal = {Communications of the ACM},
  title   = {{ImageNet classification with deep convolutional neural networks}},
  year    = {2017}
}

@inproceedings{Long2015,
  title     = {{{Fully Convolutional Networks for Semantic Segmentation}}},
  author    = {Long, Jonathan and Shelhamer, Evan and Darrell, Trevor},
  booktitle = {Proceedings of the IEEE Conference on Computer Vision and Pattern Recognition (CVPR)},
  pages     = {3431--3440},
  year      = {2015}
}

@article{Milioto2019,
  author  = {Milioto, Andres and Vizzo, Ignacio and Behley, Jens and Stachniss, Cyrill},
  journal = {IEEE International Conference on Intelligent Robots and Systems},
  title   = {{RangeNet++: Fast and Accurate LiDAR Semantic Segmentation}},
  year    = {2019}
}

@inproceedings{Moravec1989,
  author    = {Moravec, H. P.},
  editor    = {Casals, Al{\'i}cia},
  title     = {{{Sensor Fusion in Certainty Grids for Mobile Robots}}},
  booktitle = {Sensor Devices and Systems for Robotics},
  year      = {1989},
  publisher = {Springer Berlin Heidelberg},
  pages     = {253--276},
  isbn      = {978-3-642-74567-6}
}

@inproceedings{Qi2017a,
  author    = {Qi, Charles R. and Su, Hao and Mo, Kaichun and Guibas, Leonidas J.},
  booktitle = {Proceedings - 30th IEEE Conference on Computer Vision and Pattern Recognition, CVPR 2017},
  doi       = {10.1109/CVPR.2017.16},
  eprint    = {1612.00593},
  isbn      = {9781538604571},
  title     = {{PointNet: Deep learning on point sets for 3D classification and segmentation}},
  year      = {2017}
}

@inproceedings{Qi2017b,
  author    = {Qi, Charles R. and Yi, Li and Su, Hao and Guibas, Leonidas J.},
  booktitle = {Advances in Neural Information Processing Systems},
  eprint    = {1706.02413},
  issn      = {10495258},
  title     = {{PointNet++: Deep hierarchical feature learning on point sets in a metric space}},
  year      = {2017}
}

@article{Richter2019,
  author  = {Richter, Sven and Wirges, Sascha and K{\"{o}}nigshof, Hendrik and Stiller, Christoph},
  doi     = {10.1515/teme-2019-0052},
  journal = {tm - Technisches Messen},
  title   = {{Fusion of Range Measurements and Semantic Estimates in an Evidential Framework}},
  year    = {2019}
}

@inproceedings{Wirges2018Obj,
  author    = {Wirges, Sascha and Fischer, Tom and Stiller, Christoph and Frias, Jesus Balado},
  booktitle = {IEEE Conference on Intelligent Transportation Systems, Proceedings, ITSC},
  doi       = {10.1109/ITSC.2018.8569433},
  title     = {{Object Detection and Classification in Occupancy Grid Maps Using Deep Convolutional Networks}},
  year      = {2018}
}

@inproceedings{Wirges2019,
author = {Wirges, Sascha and Reith-Braun, Marcel and Lauer, Martin and Stiller, Christoph},
booktitle = {IEEE Intelligent Vehicles Symposium},
title = {{Capturing object detection uncertainty in multi-layer grid maps}},
year = {2019}
}

@inproceedings{Wu2019,
  author    = {Wu, Bichen and Zhou, Xuanyu and Zhao, Sicheng and Yue, Xiangyu and Keutzer, Kurt},
  booktitle = {Proceedings - IEEE International Conference on Robotics and Automation},
  doi       = {10.1109/ICRA.2019.8793495},
  eprint    = {1809.08495},
  isbn      = {9781538660263},
  issn      = {10504729},
  title     = {{SqueezeSegV2: Improved model structure and unsupervised domain adaptation for road-object segmentation from a LiDAR point cloud}},
  year      = {2019}
}

@inproceedings{Zhang2015,
  author    = {Zhang, Richard and Candra, Stefan A. and Vetter, Kai and Zakhor, Avideh},
  booktitle = {IEEE International Conference on Robotics and Automation},
  title     = {{Sensor fusion for semantic segmentation of urban scenes}},
  year      = {2015}
}

@article{Farabet2013,
author = {Farabet, Clement and Couprie, Camille and Najman, Laurent and Lecun, Yann},
journal = {IEEE Transactions on Pattern Analysis and Machine Intelligence},
title = {{Learning hierarchical features for scene labeling}},
year = {2013}
}

@inproceedings{Hariharan2014,
author = {Hariharan, Bharath and Arbel{\'{a}}ez, Pablo and Girshick, Ross and Malik, Jitendra},
booktitle = {Lecture Notes in Computer Science (including subseries Lecture Notes in Artificial Intelligence and Lecture Notes in Bioinformatics)},
title = {{Simultaneous detection and segmentation}},
year = {2014}
}

@inproceedings{Girshick2014,
author = {Girshick, Ross and Donahue, Jeff and Darrell, Trevor and Malik, Jitendra},
booktitle = {IEEE Computer Society Conference on Computer Vision and Pattern Recognition},
title = {{Rich feature hierarchies for accurate object detection and semantic segmentation}},
year = {2014}
}

@inproceedings{Hu2020,
author = {Hu, Qingyong and Yang, Bo and Xie, Linhai and Rosa, Stefano and Guo, Yulan and Wang, Zhihua and Trigoni, Niki and Markham, Andrew},
booktitle = {IEEE Computer Society Conference on Computer Vision and Pattern Recognition},
title = {{RandLA-Net: Efficient Semantic Segmentation of Large-Scale Point Clouds}},
year = {2020}
}

@inproceedings{Wang2015,
author = {Wang, Dominic Zeng and Posner, Ingmar},
booktitle = {Robotics: Science and Systems},
title = {{Voting for voting in online point cloud object detection}},
year = {2015}
}

@inproceedings{Zhou2018,
author = {Zhou, Yin and Tuzel, Oncel},
booktitle = {IEEE Computer Society Conference on Computer Vision and Pattern Recognition},
title = {{VoxelNet: End-to-End Learning for Point Cloud Based 3D Object Detection}},
year = {2018}
}

@inproceedings{Ohtake2003,
author = {Ohtake, Yutaka and Belyaev, Alexander and Alexa, Marc and Turk, Greg and Seidel, Hans Peter},
booktitle = {ACM SIGGRAPH 2003 Papers, SIGGRAPH '03},
doi = {10.1145/1201775.882293},
isbn = {1581137095},
title = {{Multi-level partition of unity implicits}},
year = {2003}
}

@inproceedings{Pfeiffer2011,
author = {Pfeiffer, David and Franke, Uwe},
title = {{Towards a Global Optimal Multi-Layer Stixel Representation of Dense 3D Data}},
booktitle = {Proceedings of the British Machine Vision Conference 2011},
year = {2011}
}

@article{Cordts2017,
author = {Cordts, Marius and Rehfeld, Timo and Schneider, Lukas and Pfeiffer, David and Enzweiler, Markus and Roth, Stefan and Pollefeys, Marc and Franke, Uwe},
journal = {Image and Vision Computing},
title = {{The Stixel World: A medium-level representation of traffic scenes}},
year = {2017}
}

@inproceedings{Jing2016,
author = {Jing, Huang and You, Suya},
booktitle = {International Conference on Pattern Recognition},
title = {{Point cloud labeling using 3D Convolutional Neural Network}},
year = {2016}
}

@article{Guo2020,
author = {Guo, Yulan and Wang, Hanyun and Hu, Qingyong and Liu, Hao and Liu, Li and Bennamoun, Mohammed},
journal = {IEEE Transactions on Pattern Analysis and Machine Intelligence},
title = {{Deep Learning for 3D Point Clouds: A Survey}},
year = {2020}
}

@article{Elfes1987,
author = {Elfes, Alberto},
journal = {IEEE Journal on Robotics and Automation},
title = {{Sonar-Based Real-World Mapping and Navigation}},
year = {1987}
}
